%% file: main.tex
\documentclass[a4paper, 10pt, conference]{ieeeconf}      %
\usepackage{FG2025}
\usepackage{arydshln} 
\usepackage{amsmath,amssymb,amsfonts}
\usepackage{multirow}
\usepackage{graphicx}
\usepackage{url}
\usepackage{extra_styles}
\usepackage{booktabs}
\makeatletter
\let\NAT@parse\undefined
\makeatother
\usepackage[backref=page,colorlinks=true]{hyperref}
\usepackage{cite}

\FGfinalcopy %

\IEEEoverridecommandlockouts                              %
\def\FGPaperID{165} %

\title{\LARGE \bf
Test-Time Augmentation for Pose-invariant Face Recognition
}

\author{\parbox{16cm}{\centering
   {\large Jaemin Jung$^{*}$, Youngjoon Jang$^{*}$, Joon Son Chung}\\
   {\normalsize
   Korea Advanced Institute of Science and Technology, South Korea\\}}
   \thanks{$^*$These authors contributed equally to this work.}%
}

\usepackage{fancyhdr}
\begin{document}

\ifFGfinal
\thispagestyle{empty}
\pagestyle{empty}
\else
\author{Anonymous FG2025 submission\\ Paper ID \FGPaperID \\}
\pagestyle{plain}
\fi
\maketitle

\thispagestyle{fancy}

\begin{abstract}

The goal of this paper is to enhance face recognition performance by augmenting head poses during the testing phase. Existing methods often rely on training on frontalised images or learning pose-invariant representations, yet both approaches typically require re-training and testing for each dataset, involving a substantial amount of effort. In contrast, this study proposes Pose-TTA, a novel approach that aligns faces at inference time without additional training.

To achieve this, we employ a portrait animator that transfers the source image identity into the pose of a driving image. Instead of frontalising a side-profile face -- which can introduce distortion -- Pose-TTA generates matching side-profile images for comparison, thereby reducing identity information loss. Furthermore, we propose a weighted feature aggregation strategy to address any distortions or biases arising from the synthetic data, thus enhancing the reliability of the augmented images.

Extensive experiments on diverse datasets and with various pre-trained face recognition models demonstrate that Pose-TTA consistently improves inference performance. Moreover, our method is straightforward to integrate into existing face recognition pipelines, as it requires no retraining or fine-tuning of the underlying recognition models.

\end{abstract}

\input{sec/1_introduction}

\input{sec/2_method}

\input{sec/3_experiment}
\input{sec/4_results}
\input{sec/5_conclusion}
\input{sec/6_ethic}

{\small
\bibliographystyle{ieee}
\bibliography{shortstrings, egbib}
}

\end{document}

%% file: sec/1_introduction.tex
\section{INTRODUCTION}

Face recognition has advanced considerably with the development of deep learning technologies~\cite{wang2018additive, wang2018cosface, sun2020circle, ou2024aerialface, chen2024high}. 
However, early face recognition models were trained without accounting for pose variations, leading to reduced reliability when encountering faces with unseen poses.
In response, several studies introduced datasets with diverse head poses~\cite{deng2019arcface, deng2019lightweight, yi2014learning, zhu2021webface260m, cao2018vggface2, kwak2024voxmm, atzori2024if}, laying the foundation for research on pose-agnostic face recognition.
Thanks to these studies, the scale of datasets has grown, and models~\cite{schroff2015facenet, liu2017sphereface, chen2018mobilefacenets, gao2024pointfaceformer, song2023qgface} have become more advanced, resulting in face recognition performance exceeding 90\% accuracy across various datasets.
However, recent works demonstrate that there is still room for improvement in scenarios involving diverse pose variations.
In particular, side-profile images present substantial intra-personal variations, making recognition more challenging.
Two primary approaches have been explored: (1) training models to extract pose-invariant representations~\cite{meng2021poseface,luan2023symmetrical,an2019apa} and (2) employing frontalisation techniques to synthesise images for model training~\cite{hu2018pose, qian2019unsupervised, duan2019boostgan, tran2017disentangled, banerjee2018frontalize}. 

One approach to learning pose-invariant representations is proposed in PoseFace~\cite{meng2021poseface}, where an orthogonality constraint is applied to separate identity information from head pose in the input image.
This constraint allows the model to learn representations in which images of the same identity are mapped to the same embedding space, regardless of the facial pose. 
Similarly, SSN~\cite{luan2023symmetrical} utilises a symmetrical siamese network with shared weights and contrastive loss to effectively align embeddings of the same identity across various poses. 
However, as these approaches use a pre-trained head pose extractor, their performance is inherently dependent on the accuracy of this model.

In contrast to these methods, face normalisation approaches aim to directly manipulate the facial pose to align faces into a canonical view.
CAPG-GAN~\cite{hu2018pose} performs generative model-based face frontalisation by using the target pose as input to generate faces with the desired head pose. However, converting a side-profile image to a frontal view often results in information loss, affecting features such as facial hair, wrinkles, and overall face shape.
To mitigate this issue, Dual-Attention GAN~\cite{yin2020dual} focuses on key facial features like eyes, nose, and mouth while considering facial symmetry, and High-Fidelity Face Manipulation~\cite{fu2021high} uses a high-resolution GAN with an attention mechanism to preserve facial details.
However, a major limitation of existing works is that they all require additional training processes, and validating their generalisation requires substantial computational resources and extensive experiments.

\input{figs/frame}

In this paper, we propose a method to enhance the performance of a pre-trained face recognition model during the inference process using test-time augmentation (TTA), without the need for any additional training steps. Unlike existing frameworks, we reduce the dependency on pose estimators by adapting a portrait animator~\cite{guo2024liveportrait, wang2022latent, xie2024x, deng2024portrait4d, chu2024gagavatar, Siarohin_2019_NeurIPS, siarohin2021motion, jang2023s} that takes a driving image and a source image as input, generating an image that imitates the driving pose while maintaining the identity of the source image.
Next, we focus on the issue of face distortion that occurs when converting side-profile images to frontal ones. Rather than performing face verification on a distorted frontal face, we propose a more effective approach with our method, Pose-TTA. During the pose augmentation process, instead of generating a frontal face from the unseen side of the face, Pose-TTA modifies the given face to a side profile and aligns the two images for comparison.
This method minimises the loss of identity information by comparing two spatially aligned faces, thereby providing indirect supervision for the regions of the face on which the model should focus.

Additionally, unlike traditional TTA methods~\cite{kim2020learning, shanmugam2021better, kimura2021understanding, ozturk2024intelligent, chung2018voxceleb2, shanmugam2020and} that do not use generative models and maintain object information within an image, Pose-TTA leverages a generative model to synthesise the target pose. To address potential biases and distortions in the synthetic data, we propose a weighted feature aggregation approach based on the reliability of the synthetic data. This approach ensures that the TTA process accounts for any inconsistencies in the generated data, enhancing the overall robustness and accuracy.
In particular, our method shows strong performance when there are variations in poses, demonstrating its robustness in handling non-frontal face images.
In summary, our main contributions are as follows. We introduce Pose-TTA, which enhances the performance of pre-trained face recognition models during inference without requiring additional training. To improve robustness, we propose a weighted feature aggregation that mitigates biases and distortions in synthetic data generated by the portrait animator. Finally, we demonstrate the generalisation of Pose-TTA through extensive experiments across six training datasets and five model architectures.

%% file: figs/frame.tex
\begin{figure*}[t]
\label{main_fig}
  \centering
  \vspace{-1mm}
  \includegraphics[width=0.90\linewidth]{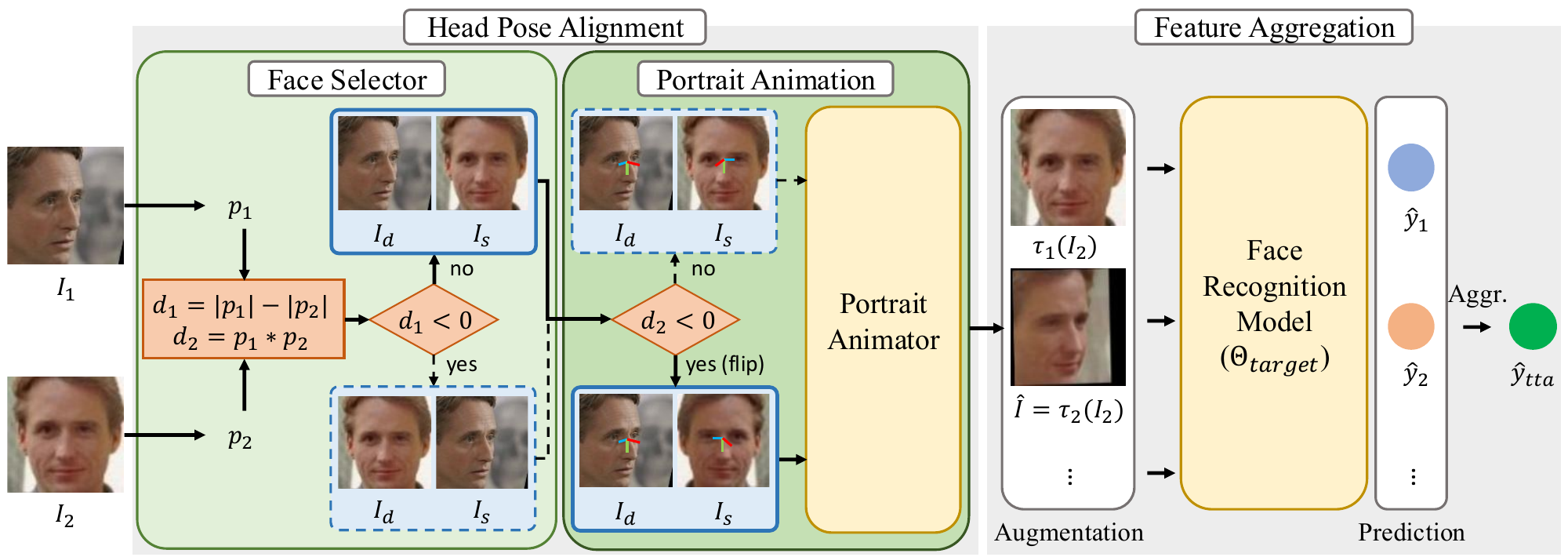}
  \vspace{-3mm}
  \caption{An overview of our Pose-TTA framework. In the head pose alignment stage, the source and driving images, $I_s$ and $I_d$, are processed through a portrait animator to generate an augmented face. In the feature aggregation stage, both the original and augmented images are passed through a pre-trained face recongition model to extract and aggregate the face embeddings $\hat{y}$ for verification, yielding the aggregated feature $\hat{y}_{tta}$.
  Aggr. denotes aggregation.
  }
  \label{fig:framework}
  \vspace{-5mm}
\end{figure*}

%% file: sec/2_method.tex
\section{PROPOSED METHOD}
The overall framework of the proposed Pose-TTA is shown in~\Fref{fig:framework}. Our framework consists of two main stages: head pose alignment and feature aggregation.
In the head pose alignment stage, the Face Selector takes two face images, $I_{1}$ and $I_{2}$, as inputs and determines the source image $I_{s}$, which retains the identity, and the driving image $I_{d}$, which provides the target pose. Then, $I_{s}$ and $I_{d}$ are passed through a pre-trained portrait animator to generate an augmented face $\hat{I}$. In the Feature Aggregation stage, both the original and multiple augmented images are processed through a pre-trained face recognition model to extract the face embedding $\hat{y}$. Finally, the aggregated feature $\hat{y}_{tta}$, a weighted combination of the augmented and original features, is used for verification.

\input{tab/sota}
\subsection{Head Pose Alignment}
\newpara{Face selector.} 
The Face Selector is responsible for selecting the most suitable face pose for transformation in order to minimise identity loss and distortion.
A critical issue in face frontalisation is that synthesising a frontal view from a partially occluded face can lead to facial information distortion.
The Face Selector takes two images as input and utilises an off-the-shelf head pose extractor~\cite{ruiz2018fine} to estimate yaw, which is a key factor influencing facial information loss. Based on this, the image with the smaller absolute yaw value ($p_2$) is selected as the source image $I_{s}$, while the image with the larger absolute yaw value ($p_1$) is designated as the driving image $I_{d}$. 
Note that the off-the-shelf head pose estimator is not used during the pose augmentation process, thereby eliminating dependence on its performance.

\newpara{Portrait animation.}
The goal of portrait animation is to transform the source image $I_{s}$ into an image $\hat{I}$ that replicates the head pose of the driving image $I_{d}$.
For this process, we employ LivePortrait~\cite{guo2024liveportrait} as the portrait animator. The key reasons for this choice are: (1) comparable performance to diffusion-based methods~\cite{zeng2023face, han2024face, wei2024aniportrait} while achieving faster inference speed; (2) independent control of facial expressions and head pose; and (3) unlike methods that require explicit roll/pitch/yaw inputs, LivePortrait conditions the transformation directly on a driving image, eliminating dependence on the head pose estimator's accuracy.
However, directly inputting $I_{s}$ and $I_{d}$ into the portrait animator can lead to unexpected distortions when there is a substantial difference in head pose between the two images.
This happens because the model must generate the occluded side of the face from $I_{s}$, which can introduce artifacts.

To mitigate this issue, we leverage facial symmetry. Using the yaw values extracted by the Face Selector, we identify cases where $I_{s}$ and $I_{d}$ have opposite facial directions (i.e., when the product of their yaw values is negative). In such cases, we horizontally flip $I_{s}$ to roughly align its face direction with $I_{d}$, reducing the discrepancy between the two images.  
Furthermore, we experimentally observe that if the source image also mimics the facial expression of the driving image, it biases features such as eye shape, mouth shape, and facial contours toward those of the driving image. Therefore, we perform augmentation only on head pose.
Finally, we obtain the augmented image $\hat{I}$ with minimal distortion from the pose transformation process.

\subsection{Test Time Adaptation}
If a candidate set of augmentations $\mathcal{T} = \{\tau_1, \tau_2, \dots, \tau_{|\mathcal{T}|}\}$ is selected at test time,  conventional test-time augmentation can be formulated as:
\begin{equation}
\hat{y}_{tta} = \frac{1}{|\mathcal{T}|} \sum_{i=1}^{|\mathcal{T}|} \Theta_{\text{target}}(\tau_i(x)),
\end{equation}
where $\Theta_{target}$ is a neural network trained on the target dataset, and $x$ is an input image. 
In our case, $\Theta_{target}$ represents the pre-trained face recognition model. The set $\tau_i(x)$ consists of four representations when the Face Selector designates an image as the source: the original image, the augmented image, the horizontally flipped original image, and the flipped augmented image, incorporating facial symmetry. When an image is designated as the driving image by the Face Selector, $\tau_i(x)$ includes only two representations: the original image and the flipped original image.

\newpara{Feature aggregation.}
\cite{shanmugam2021better} observed that simply averaging predictions from TTA-transformed images can sometimes degrade accuracy by turning correct predictions into incorrect ones. Based on this, we introduce a weighting mechanism into the TTA process, assigning lower weights to synthetic images due to potential distortions, thereby mitigating negative effects while preserving the benefits of TTA. The weighted feature aggregation is defined by the equation:

\begin{equation}
\hat{y}_{tta} = \frac{1}{|\mathcal{T}|} \sum_{i=1}^{|\mathcal{T}|} w_{\mathcal{T}}*\Theta_{\text{target}}(\tau_i(x)).
\end{equation}

Here, $w_{\mathcal{T}}$ is determined by the following conditions:

\begin{equation}
w_{\mathcal{T}} =
\begin{cases}
    w_{syn}, & \text{if } \tau_i(x) \text{ is synthetic data} \\
    w_{real}, & \text{if } \tau_i(x) \text{ is real data}
\end{cases}
\end{equation}

The values of $w_{\text{syn}}$ and $w_{\text{real}}$ are hyperparameters for balancing the embeddings, and in this paper, they are set to 0.25 and 0.75, respectively. Since the feature scale of $\hat{y}_{\text{tta}}$ can vary depending on the number of elements in $\tau_i(x)$, $\hat{y}_{\text{tta}}$ is ultimately normalised along the channel axis before face verification.

%% file: tab/sota.tex
\begin{table*}[!t]
\centering
\vspace{-1mm}
\caption{Effectiveness of Pose-TTA. TTA refers to test-time augmentation. Baseline results without TTA use the aggregation of original and flipped embeddings. All experiments use $w_{real} = 0.75$. $\dag$ indicates our trained models.}
\vspace{-3mm}
\footnotesize
\renewcommand{\arraystretch}{0.5}
\label{tab:sota}
\scalebox{0.85}{
\begin{tabular}{lcccccccccc}
\toprule
\multicolumn{1}{c}{\multirow{2}{*}{Method}} & \multicolumn{1}{c}{\multirow{2}{*}{Backbone}} & \multicolumn{1}{c}{\multirow{2}{*}{Train Data}} & \multicolumn{1}{c}{\multirow{2}{*}{TTA}}  &\multicolumn{6}{c}{Accuracy (\%)}  \\
   &  &  &  & CPLFW~\cite{zheng2018cross} &  CFP-FP~\cite{sengupta2016frontal} &  LFW~\cite{huang2008labeled} &  CALFW~\cite{zheng2017cross}  &  AgeDB~\cite{moschoglou2017agedb}  &  Avg  \\ \midrule
\multicolumn{1}{c}{\multirow{23}{*}{AdaFace~\cite{kim2022adaface}}}  &   \multicolumn{1}{c}{\multirow{23}{*}{ResNet101}}    & \multicolumn{1}{c}{\multirow{2}{*}{WebFace12M~\cite{zhu2021webface260m}}}  & \xmark  &  94.57 & 99.24 & 99.82 & 96.12 & 98.00 & 97.55    \\
  &       &   & \checkmark  & \hspace{16pt} 95.40 {\scriptsize \blue{+0.83}} & \hspace{15pt} 99.23 {\scriptsize \red{-0.01}} & \hspace{16pt} 99.83 {\scriptsize \blue{+0.01}} & \hspace{15pt} 96.08 {\scriptsize \red{-0.04}} & \hspace{15pt} 97.87 {\scriptsize \red{-0.03}} & \bf{97.68}    \\ \cmidrule{3-10} 
  &       & \multicolumn{1}{c}{\multirow{2}{*}{WebFace4M~\cite{zhu2021webface260m}}}  & \xmark   & 94.63 & 99.27 & 99.83 & 96.05 & 97.90 & 97.54    \\
  &       &   & \checkmark   & \hspace{16pt} 95.28 {\scriptsize \blue{+0.65}} & \hspace{15pt} 99.24 {\scriptsize \red{-0.03}} & \hspace{15pt} 99.80 {\scriptsize \red{-0.03}} & \hspace{16pt} 96.10 {\scriptsize \blue{+0.05}} & \hspace{16pt} 98.02 {\scriptsize \blue{+0.12}} & \bf{97.69}    \\ \cmidrule{3-10} 
  &       & \multicolumn{1}{c}{\multirow{2}{*}{MS1MV3~\cite{deng2019lightweight}}}   & \xmark   & 93.92 & 99.09 & 99.83 & 96.02 & 98.18 & 97.41    \\
  &       &    & \checkmark   & \hspace{16pt} 94.75 {\scriptsize \blue{+0.83}} & \hspace{16pt} 99.16 {\scriptsize \blue{+0.07}} & 99.83 & \hspace{16pt} 96.20 {\scriptsize \blue{+0.18}} & \hspace{16pt} 98.27 {\scriptsize \blue{+0.09}} & \bf{97.64}    \\ \cmidrule{3-10} 
  &       & \multicolumn{1}{c}{\multirow{2}{*}{MS1MV2~\cite{deng2019arcface}}}   & \xmark   & 93.53 & 98.67 & 99.80 & 96.12 & 98.05 & 97.23    \\
  &       &    & \checkmark   & \hspace{16pt} 94.28 {\scriptsize \blue{+0.75}} & 98.67 & \hspace{16pt} 99.82 {\scriptsize \blue{+0.02}} & \hspace{15pt} 96.10 {\scriptsize \red{-0.02}} & 98.05 & \bf{97.38}    \\ \cmidrule{3-10} 
  &       & \multicolumn{1}{c}{\multirow{2}{*}{CASIA-Webface$^{\dag}$~\cite{yi2014learning}}}   & \xmark   & 90.12 & 97.37 & 99.32 & 93.60 & 94.88 & 95.06    \\
  &       &   & \checkmark   & \hspace{16pt} 90.17 {\scriptsize \blue{+0.05}} & \hspace{16pt} 97.49 {\scriptsize \blue{+0.12}} & 99.32 & 93.60 & \hspace{16pt} 95.05 {\scriptsize \blue{+0.17}} & \bf{95.13}    \\ \cmidrule{3-10} 
  &       & \multicolumn{1}{c}{\multirow{2}{*}{DCFace$^{\dag}$~\cite{kim2023dcface}}}   & \xmark   & 82.98 & 91.37 & 98.58 & 91.90 & 90.65 & 91.10    \\
  &       &   & \checkmark   & \hspace{16pt} 84.97 {\scriptsize \blue{+1.99}} & \hspace{16pt} 92.10 {\scriptsize \blue{+0.73}} & 98.58 & \hspace{15pt} 91.82 {\scriptsize \red{-0.08}} & \hspace{16pt} 90.78 {\scriptsize \blue{+0.13}} & \bf{91.65}    \\
  \midrule
\multicolumn{1}{c}{\multirow{19}{*}{AdaFace~\cite{kim2022adaface}}}  &   \multicolumn{1}{c}{\multirow{2}{*}{ViT}}    & \multicolumn{1}{c}{\multirow{2}{*}{WebFace4M~\cite{zhu2021webface260m}}}  & \xmark   & 94.97 & 98.94 & 99.80 & 96.03 & 97.48 & 97.44    \\
  &       &  & \checkmark   & \hspace{16pt} 95.08 {\scriptsize \blue{+0.09}} & \hspace{16pt} 99.10 {\scriptsize \blue{+0.16}} & \hspace{16pt} 99.82 {\scriptsize \blue{+0.02}} & 96.03 & \hspace{15pt} 97.22 {\scriptsize \red{-0.26}} & \bf{97.45}    \\ \cmidrule{2-10} 
  &   \multicolumn{1}{c}{\multirow{7}{*}{ResNet50}}        & \multicolumn{1}{c}{\multirow{2}{*}{WebFace4M~\cite{zhu2021webface260m}}}  & \xmark   & 94.17 & 98.99 & 99.78 & 95.98 & 97.78 & 97.34    \\
  &       &   & \checkmark   & \hspace{16pt} 94.83 {\scriptsize \blue{+0.66}} & \hspace{15pt} 98.89 {\scriptsize \red{-0.10}} & \hspace{16pt} 99.82 {\scriptsize \blue{+0.04}} & \hspace{15pt} 95.95 {\scriptsize \red{-0.03}} & \hspace{15pt} 97.73 {\scriptsize \red{-0.05}} & \bf{97.44}    \\ \cmidrule{3-10} 
  &       & \multicolumn{1}{c}{\multirow{2}{*}{CASIA-Webface~\cite{yi2014learning}}} & \xmark   & 90.02 & 97.04 & 99.37 & 93.43 & 94.40 & 94.85    \\
  &       &  & \checkmark   & \hspace{16pt} 90.35 {\scriptsize \blue{+0.32}} & \hspace{16pt} 97.06 {\scriptsize \blue{+0.02}} & \hspace{15pt} 99.30 {\scriptsize \red{-0.07}} & \hspace{16pt} 93.52 {\scriptsize \blue{+0.09}} & \hspace{15pt} 94.17 {\scriptsize \red{-0.23}} & \bf{94.88}   \\ \cmidrule{2-10} 
  &   \multicolumn{1}{c}{\multirow{7}{*}{ResNet18}}    & \multicolumn{1}{c}{\multirow{2}{*}{WebFace4M~\cite{zhu2021webface260m}}}  & \xmark   & 92.28 & 97.80 & 99.58 & 95.52 & 96.48 & 96.33    \\
  &      &   & \checkmark   & \hspace{16pt} 92.85 {\scriptsize \blue{+0.57}} & \hspace{16pt} 97.91 {\scriptsize \blue{+0.11}} & 99.58 & \hspace{15pt} 95.50 {\scriptsize \red{-0.02}} & \hspace{15pt} 96.35 {\scriptsize \red{-0.13}} & \bf{96.44}    \\ \cmidrule{3-10} 
  &       & \multicolumn{1}{c}{\multirow{2}{*}{CASIA-Webface~\cite{yi2014learning}}} & \xmark   & 87.00 & 94.81 & 99.22 & 92.65 & 92.68 & 93.27    \\
  &       &  & \checkmark   & \hspace{16pt} 87.87 {\scriptsize \blue{+0.87}} & \hspace{16pt} 95.34 {\scriptsize \blue{+0.53}} & \hspace{16pt} 99.32 {\scriptsize \blue{+0.10}} & \hspace{16pt} 92.97 {\scriptsize \blue{+0.32}} & 92.68 & \bf{93.64}    \\ 
  \midrule
\multicolumn{1}{c}{\multirow{11}{*}{ArcFace~\cite{deng2019arcface}}}  &   \multicolumn{1}{c}{\multirow{11}{*}{ResNet101}}    & \multicolumn{1}{c}{\multirow{2}{*}{WebFace4M~\cite{zhu2021webface260m}}}  & \xmark  & 94.35 & 99.21 & 99.78 & 96.00 & 97.95 & 97.46    \\
  &       &   & \checkmark  & \hspace{16pt} 95.02 {\scriptsize \blue{+0.67}} & \hspace{16pt} 99.24 {\scriptsize \blue{+0.03}} & 99.78 & \hspace{15pt} 95.88 {\scriptsize \red{-0.12}} & \hspace{16pt} 98.03 {\scriptsize \blue{+0.08}} & \bf{97.59}    \\ \cmidrule{3-10} 
  &       & \multicolumn{1}{c}{\multirow{2}{*}{CASIA-Webface$^{\dag}$~\cite{yi2014learning}}}   & \xmark   & 89.98 & 97.23 & 99.47 & 93.52 & 94.63 & 94,97    \\
  &       &   & \checkmark   & \hspace{16pt} 90.08 {\scriptsize \blue{+0.10}} & \hspace{16pt} 97.53 {\scriptsize \blue{+0.30}} & \hspace{15pt} 99.38 {\scriptsize \red{-0.09}} & \hspace{16pt} 93.63 {\scriptsize \blue{+0.11}} & \hspace{15pt} 94.32 {\scriptsize \red{-0.31}} & \bf{94.99}    \\ \cmidrule{3-10} 
  &       & \multicolumn{1}{c}{\multirow{2}{*}{DCFace$^{\dag}$~\cite{kim2023dcface}}}   & \xmark   & 82.50 & 90.49 & 98.68 & 92.05 & 90.83 & 90.91    \\
  &       &   & \checkmark   & \hspace{16pt} 84.22 {\scriptsize \blue{+1.72}} & \hspace{16pt} 91.27 {\scriptsize \blue{+0.78}} & \hspace{15pt} 98.65 {\scriptsize \red{-0.03}} & \hspace{15pt} 92.03 {\scriptsize \red{-0.02}} & \hspace{15pt} 90.70 {\scriptsize \red{-0.07}} & \bf{91.37} \\ 
\bottomrule
\end{tabular}
}
\vspace{-4mm}
\end{table*}

%% file: sec/3_experiment.tex
\section{EXPERIMENTS}

\subsection{Experimental setup}

\newpara{Baselines.}
To validate the generalisation capability of the proposed Pose-TTA method, we select pre-trained models trained with a combination of various backbone models (ViT~\cite{dosovitskiy2020vit}, ResNet18, ResNet50, ResNet101~\cite{he2016deep}) and two training losses (AdaFace~\cite{kim2022adaface}, ArcFace~\cite{deng2019arcface}).
The datasets used in the experiments are as follows: Casia-WebFace~\cite{yi2014learning}, MS1MV2~\cite{deng2019arcface}, MS1MV3~\cite{deng2019lightweight}, WebFace4M, WebFace12M~\cite{zhu2021webface260m}, and the synthetic dataset DCFace~\cite{kim2023dcface}. These datasets contain 0.49M, 5.8M, 5.1M, 4.2M, 12M, and 0.5M facial images, respectively.

\newpara{Dataset and evaluation.}
We evaluate the face recognition models on five datasets: CPLFW~\cite{zheng2018cross}, 
CFP-FP~\cite{sengupta2016frontal}, LFW~\cite{huang2008labeled}, 
CALFW~\cite{zheng2017cross}, and AgeDB~\cite{moschoglou2017agedb}.
CPLFW includes large pose variations, while CFP-FP consists of pairs of frontal and 90-degree side-profile faces. LFW contains variations in lighting and expression, with most images being frontal or near-frontal~\cite{an2019apa}. CALFW and AgeDB-30 feature images of the same person at different ages.
We follow the evaluation protocol of CVLFace~\cite{cvlface}

%% file: sec/4_results.tex
\section{RESULTS}

\subsection{Effectiveness of Pose-TTA}
\Tref{tab:sota} presents the performance of face recognition models across multiple datasets, demonstrating the improvement achieved by applying our Pose-TTA during inference.
The results show that applying TTA consistently improves performance on CPLFW and CFP-FP, with average accuracy gains of 0.72\% and 0.19\% across all models, respectively . Notably, these datasets contain significant head pose variations, suggesting that our augmentation strategy effectively mitigates challenges from extreme poses and enhances model robustness in unconstrained face verification scenarios.

Conversely, for datasets dominated by frontal faces, such as LFW and CALFW, the performance change remains marginal. In some cases, we observe a slight drop in accuracy (0.04\% on AgeDB-30), likely due to the introduction of unnecessary variations that do not contribute meaningful information to these datasets. These findings suggest that our method can enhance face verification performance in real-world scenarios where diverse head poses are encountered, while maintaining stable performance even when pose variations are minimal.

\subsection{Comparison with Frontalisation.}
As mentioned earlier, to quantitatively analyse the issue of facial distortion when using frontalisation in TTA, we compare our method with the approach presented in~\Tref{tab:mixed}. Notably, for datasets with diverse head poses (CPLFW, CFP-FP), frontalisation leads to a significant performance drop, resulting in an average degradation of 0.12\% compared to the baseline model without TTA. In contrast, our method, which performs pose augmentation while minimising identity distortion, achieves an average performance improvement of 0.37\%. Furthermore, when face direction alignment using flipping is omitted (denoted as `Ours w/o flip'), by comparing the yaw angles of the two faces before feeding them into the portrait animator, we observe an average performance drop of 0.04\%. This demonstrates that simple alignment using flipping, which leverages facial symmetry to compensate for large pose differences, effectively reduces identity distortion during the portrait animation process.

\input{tab/mixed}

\input{tab/weight}

\input{figs/fail}
\subsection{Effectiveness of Weighted Feature Aggregation.}
Unlike traditional TTA methods, which apply transformations such as translation and rotation while preserving the original structure of objects within the image, our approach generates synthetic data by modifying the face within the image. Consequently, biases inherent in the generative model and subtle distortions introduced during the generation process are unavoidable. To ensure reliable feature extraction when using both real and synthetic data, we propose a weighted aggregation method and present an ablation study on the weighting strategy in~\Tref{tab:weight}. The results indicate that performance decreases when dependence on real data is reduced. Notably, when the weight assigned to real data is increased to 0.75, our method outperforms the conventional approach of averaging features from original and augmented data (i.e., $w_{real}$ = 0.5). This demonstrates that our proposed method effectively mitigates the drawbacks of synthetic data, such as distortion and bias, ultimately contributing to improved performance.

\subsection{Qualitative Comparison}
We present the quality of head pose augmentation achieved by the proposed method in~\Fref{fig:failure}. As we pointed out, frontalisation methods inevitably generate unseen regions in the source image, leading to distortions in the original identity. Furthermore, when there is a significant difference in head pose between the source and driving faces, distortions occur during the portrait animation process (see the `w/o flip' case). Consequently, by flipping the source image based on the yaw value obtained from the Face Selector and applying portrait transformation, we can minimise distortions from the original image and better preserve the identity.

%% file: tab/mixed.tex
\begin{table}[!t]
\centering
\caption{
Comparison of pose augmentation methods using a ResNet18 model trained on CASIA-Webface with AdaFace.}
\vspace{-2mm}
\footnotesize
\renewcommand{\arraystretch}{0.8}
\label{tab:mixed}
\scalebox{0.85}{
\begin{tabular}{lccccccccc}
\toprule
\multicolumn{1}{l}{\multirow{2}{*}{Method}}  &\multicolumn{6}{c}{Accuracy (\%)}  \\ 
    & CPLFW &  CFP-FP &  LFW &  CALFW  &  AgeDB  &  Avg  \\ \midrule
Baseline        & 87.00 & 94.81 & 99.22 & 92.65  & 92.68 & 93.27         \\
Frontalisation  & 86.83 & 94.41 & 99.23 & 92.65 & 92.65 & 93.15         \\
Ours w/o flip    & 87.82 & 95.30 & \bf{99.32} & 92.83 & \bf{92.72} & 93.60         \\
Ours            & \bf{87.87} & \bf{95.34} & \bf{99.32} & \bf{92.97} & 92.68 & \bf{93.64}         \\
\bottomrule
\end{tabular}
}

\end{table}

%% file: tab/weight.tex
\begin{table}[!t]
\centering
\caption{Performance comparison based on weight in Weighted Feature Aggregation  using a ResNet18 model trained on CASIA-Webface with AdaFace.}
\vspace{-2mm}
\footnotesize
\renewcommand{\arraystretch}{0.8}
\label{tab:weight}
\scalebox{0.85}{
\begin{tabular}{ccccccccccc}
\toprule
\multicolumn{2}{c}{Aggr. Weight}   & \multicolumn{6}{c}{Accuracy (\%)}  \\ 
\multirow{1}{*}{$w_{\text{real}}$}    & \multirow{1}{*}{$w_{\text{syn}}$} & CPLFW &  CFP-FP &  LFW~ &  CALFW  &  AgeDB  &  Avg  \\ \midrule
0.00   & 1.00    & 85.77 & 90.07 & 98.92 & 91.72 & 91.50 & 91.60         \\
0.25   & 0.75    & 87.23 & 93.49 & 99.15 & 92.37 & 92.47 & 92.94         \\
0.50   & 0.50    & \bf{87.88} & 94.84 & 99.23 & 92.92 & \bf{92.75} & 93.52         \\
0.75   & 0.25    & 87.87 & \bf{95.34} & \bf{99.32} & \bf{92.97} & 92.68 & \bf{93.64}         \\
1.00   & 0.00    & 87.00 & 94.81 & 99.22 & 92.65 & 92.68 & 93.27         \\
\bottomrule
\end{tabular}
}  
\vspace{-2mm}

\end{table}

%% file: figs/fail.tex
\begin{figure}[t]
  \centering
  \includegraphics[width=0.99\linewidth]{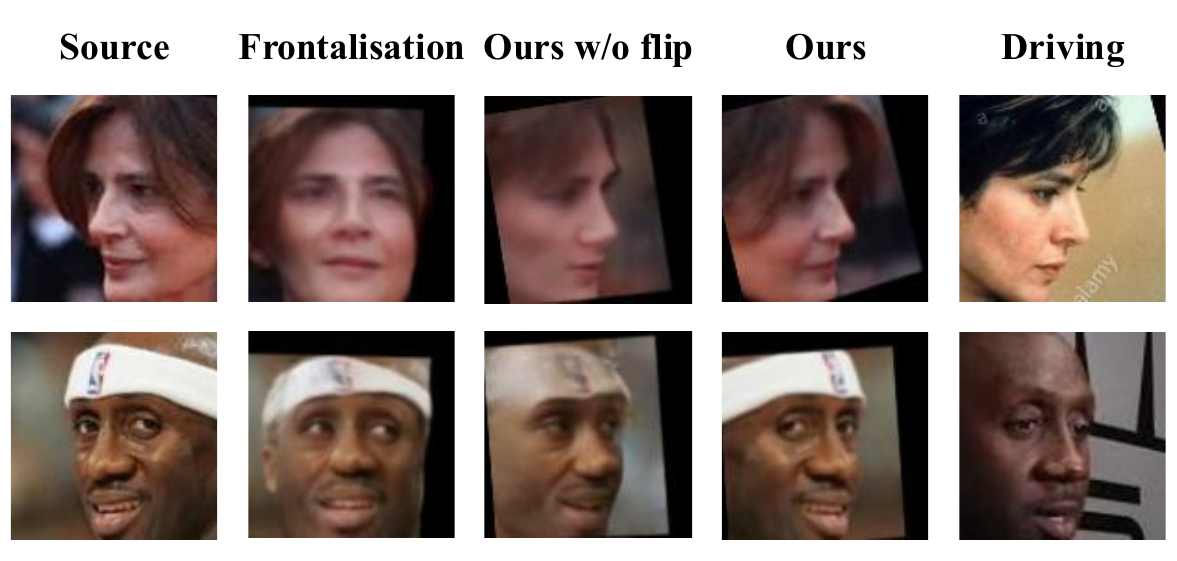}
  \vspace{-8mm}
  \caption{
  Head pose augmentation quality comparison. The proposed method minimises identity distortions by flipping the source image and applying portrait transformation.
  }
  \vspace{-1mm}
  \label{fig:failure}
\end{figure}

%% file: sec/5_conclusion.tex
\section{CONCLUSION}
In this paper, we present Pose-TTA, a novel approach that enhances pre-trained face recognition models during inference by utilising test-time augmentation with a portrait animator. Unlike traditional methods that rely on frontalisation, our approach aligns side-profile images, minimising identity distortion and improving verification accuracy. Additionally, our weighted feature aggregation strategy effectively addresses biases in synthetic data, ensuring a more reliable augmentation process. Extensive experiments across six datasets and five face recognition frameworks demonstrate the effectiveness and generalisation of Pose-TTA, showcasing its superiority over conventional TTA and face frontalisation methods. This approach offers an efficient way to improve face recognition models without extensive retraining and shows the potential to be applied to other face-related tasks in complex environments.

\section{ACKNOWLEDGMENTS}
This work was supported by HDC Labs. We would like to thank Chansung Jung, Gunhee Lee and Chongmin Park for helpful discussions.

%% file: sec/6_ethic.tex
\section*{ETHICAL IMPACT STATEMENT}

\newpara{Potential Risks.}  
We introduce Pose-TTA, a test-time augmentation method designed to enhance pose-invariant face recognition. While the method improves accuracy by proposing pose-agnostic TTA framework, it also raises ethical concerns related to fairness, privacy, and potential misuse.  

One major risk lies in the inherent biases present in publicly available face recognition datasets~\cite{deng2019arcface, deng2019lightweight, yi2014learning, zhu2021webface260m}. These datasets often exhibit demographic imbalances, leading to variations in recognition accuracy across different ethnicities, genders, and age groups~\cite{buolamwini2018gender, leslie2020understanding, sixta2020fairface, mehrabi2021survey}. If these biases are not carefully addressed, the proposed method may cause performance disparities across different demographic groups.
Additionally, large-scale face recognition datasets have raised serious ethical concerns, particularly regarding privacy violations and the lack of informed consent~\cite{bae2023digiface}. Many widely used datasets~\cite{deng2019arcface, zhu2021webface260m} have been constructed by indiscriminately collecting web images, often without the subjects' approval. In some cases, the classification of ``celebrities'' has been broadly applied to individuals with an online presence, raising further ethical concerns. As a result of these issues, public access to several of these datasets has been revoked~\cite{van2020ethical}, emphasising the need for stricter data governance. These concerns underscore the necessity of ethical data collection practices and adherence to informed consent principles in face recognition research.  

Moreover, recent advances in generative models across various domains—such as image and audio generation~\cite{tian2024visual, li2024autoregressive, wang2023neural, jung2025voicedit}—have significantly expanded both the capabilities and the potential risks of AI-driven systems. In particular, the ability to synthetically manipulate head poses raises concerns about identity spoofing or potential misuse in unauthorised applications, such as deepfake generation~\cite{tolosana2020deepfakes}. Although our Pose-TTA does not explicitly involve identity synthesis, any system that modifies facial representations must consider risks related to authenticity and trustworthiness in biometric authentication.

\newpara{Risk-Mitigation Strategies.}
To mitigate the aforementioned risks associated with dataset bias, privacy concerns, and potential misuse, we take several precautions in the design and implementation of Pose-TTA. 

First, we analyse Pose-TTA’s performance across multiple datasets~\cite{deng2019arcface, deng2019lightweight, yi2014learning, zhu2021webface260m} and models to identify potential biases in recognition accuracy across demographic groups. This approach helps ensure that our augmentation technique does not disproportionately favour specific populations.
Additionally, our method operates solely during inference as a test-time augmentation approach, without modifying the training process. This design minimises the risk of amplifying dataset biases, as the underlying recognition models remain unchanged. 
Furthermore, to address privacy concerns, we evaluated Pose-TTA using synthetic face datasets~\cite{kim2023dcface}, thereby reducing reliance on sensitive real-world biometric data. This approach helps mitigate privacy risks related to data misuse and consent, while enabling robust model validation in privacy-preserving environments.
Lastly, we promote the responsible use of Pose-TTA by restricting code access to organisations that agree to ethical usage guidelines. We emphasise its intended applications in research and authentication, while actively discouraging its use in surveillance, unauthorised identity manipulation, or adversarial AI applications.

\newpara{Benefit-Risk Analysis.}
Despite the potential risks, Pose-TTA offers substantial benefits when applied responsibly. By addressing pose variations in face recognition, it improves model robustness without requiring additional training, making it an efficient and scalable solution for real-world applications. This improvement is particularly valuable for identity verification systems, where extreme pose variations often lead to recognition failures. Additionally, our test-time augmentation approach provides a computationally efficient alternative to traditional data augmentation methods, reducing the need for extensive retraining on pose-augmented datasets.
Moreover, Pose-TTA maintains high performance even when applied to synthetic face datasets, underscoring its suitability for privacy-conscious applications. This demonstrates its ability to deliver accurate results while reducing dependence on sensitive personal data, thereby aligning with ethical AI principles and supporting privacy-preserving use cases. However, we acknowledge that face recognition technology remains ethically complex, and we strongly advocate for continued scrutiny, fairness evaluations, and regulatory oversight in its deployment. By ensuring transparency, responsible usage, and ongoing assessment of its societal impact, Pose-TTA can contribute positively to the advancement of ethical and reliable face recognition systems.